\documentclass[twocolumn, fontsize=12pt]{article}
\usepackage[margin=0.70in]{geometry}
\usepackage{lipsum,mwe,abstract}
\usepackage[T1]{fontenc} 
\usepackage[english]{babel} 
\usepackage{glossaries}
\usepackage[table,xcdraw]{xcolor}
\usepackage{fancyhdr} % Custom headers and footers
\usepackage{lipsum}
\usepackage{multicol}
\usepackage{amsmath,amsfonts,amsthm} % Math packages
\usepackage{amsmath}
\usepackage{mathrsfs}
\usepackage{wrapfig}
\usepackage{graphicx}
\usepackage{float}
\usepackage{subcaption}
\usepackage{comment}
\usepackage{enumitem}
\usepackage{hyperref}
\usepackage{cuted}
\usepackage{placeins}
\usepackage{sectsty} % Allows customizing section commands
\allsectionsfont{\normalfont \normalsize \scshape} % Section names in small caps and normal fonts

\renewenvironment{abstract} % Change how the abstract look to remove margins
 {\small
  \begin{center}
  \bfseries \abstractname\vspace{-.5em}\vspace{0pt}
  \end{center}
  \list{}{%
    \setlength{\leftmargin}{0mm}
    \setlength{\rightmargin}{\leftmargin}%
  }
  \item\relax}
 {\endlist}
 
\makeatletter
\renewcommand{\maketitle}{\bgroup\setlength{\parindent}{0pt} % Change how the title looks like
\begin{flushleft}
  \textbf{\@title}
  \@author \\ 
  \@date
\end{flushleft}\egroup
}
\makeatother

\title{
\Large 
Review of Zero-Shot and Few-Shot AI Algorithms in The Medical Domain
\\
[12pt] 
}
\date{\today}
\author{Maged Badawi  \textsuperscript{1} - maged.badawi@fau.de, Mohammedyahia Abushanab\textsuperscript{1} - moh.y.abushanab@fau.de, Sheethal Bhat \textsuperscript{1} - sheethal.bhat@fau.de, Andreas Maier \textsuperscript{1} - andreas.maier@fau.de.} % Authors

\begin{document}

\twocolumn[ 
  \maketitle
  \begin{multicols}{2}
    \textbf{} 

    \columnbreak

    \textbf{}   
  \end{multicols}
]

% --------------- ABSTRACT
\begin{abstract}
   In this paper, different techniques of few-shot, zero-shot, and regular object detection have been investigated. The need for few-shot learning and zero-shot learning techniques is crucial and arises from the limitations and challenges in traditional machine learning, deep learning, and computer vision methods where they require large amounts of data, plus the poor generalization of those traditional methods. 
     Those techniques can give us prominent results by using only a few training sets reducing the required amounts of data and improving the generalization.
     
     This survey will highlight the recent papers of the last three years that introduce the usage of few-shot learning and zero-shot learning techniques in addressing the challenges mentioned earlier. In this paper we reviewed the Zero-shot, few-shot and regular object detection methods and categorized them in an understandable manner. Based on the comparison made within each category. It been found that the approaches are quite impressive. 
     This integrated review of diverse papers on few-shot, zero-shot, and regular object detection reveals a shared focus on advancing the field through novel frameworks and techniques. A noteworthy observation is the scarcity of detailed discussions regarding the difficulties encountered during the development phase. Contributions include the introduction of innovative models, such as ZSD-YOLO and GTNet, often showcasing improvements with various metrics such as mean average precision (mAP),Recall@100 (RE@100), the area under the receiver operating characteristic curve (AUROC) and precision. These findings underscore a collective move towards leveraging vision-language models for versatile applications, with potential areas for future research including a more thorough exploration of limitations and domain-specific adaptations.

\end{abstract}
\rule{\linewidth}{0.5pt}

% --------------- MAIN CONTENT

% If you want to put a big figure over two columns, you can use \begin{figure*}
% The figure will be sent on top of the next page 
% \begin{figure*}   
%     \centering
%     \includegraphics[width=0.9\textwidth]{Dummy_fig.png}
%     \caption{Caption}
% \end{figure*}

% You can also use the following command, but it can be more tricky to place the figure exactly where you want
% \begin{strip}
%     \centering\noindent
%     \includegraphics[width=0.9\textwidth]{Dummy_fig.png}
%     \captionof{figure}{Caption}
% \end{strip}

\section{Introduction} 
In the continually evolving domains of machine learning where computers learn patterns from data to make predictions or decisions without explicit programming. Computer vision describes the use of computer science in enabling machines to interpret visual information to perform tasks like object detection along side deep learning where deep neural networks are used to automatically learn complex patterns and features from the data. The field of object detection, where a computer vision task involves in identifying and locating objects within images or videos, has witnessed remarkable advancements.  Traditional object detection models have historically relied on substantial volumes of precisely annotated data for their training, a requirement that can prove to be quite challenging when dealing with numerous objects and diverse real-world scenarios. Moreover, these models suffer from the task of adapting and accurately recognizing entirely new categories of objects, which can pose a significant obstacle when deploying them in practical applications.

For the previous reasons and to unlock new possibilities, the aspects of first, \textbf{few-shot} that denotes a machine learning paradigm that involves training a model to make accurate predictions with only a small number of instances per class, \textbf{zero-shot} that is similar to few-shot paradigm but with the difference that it performs tasks on classes that have never seen during the training phase, and \textbf{regular object detection} have emerged as promising solutions. These innovative approaches hold the potential to reshape various domains, from computer vision and segmentation as You-Only-Look-Once (YOLO), the real-time fast object detection model, to medical diagnosis and beyond.

\begin{figure}[h]
    \centering
    \includegraphics[width=1\linewidth]{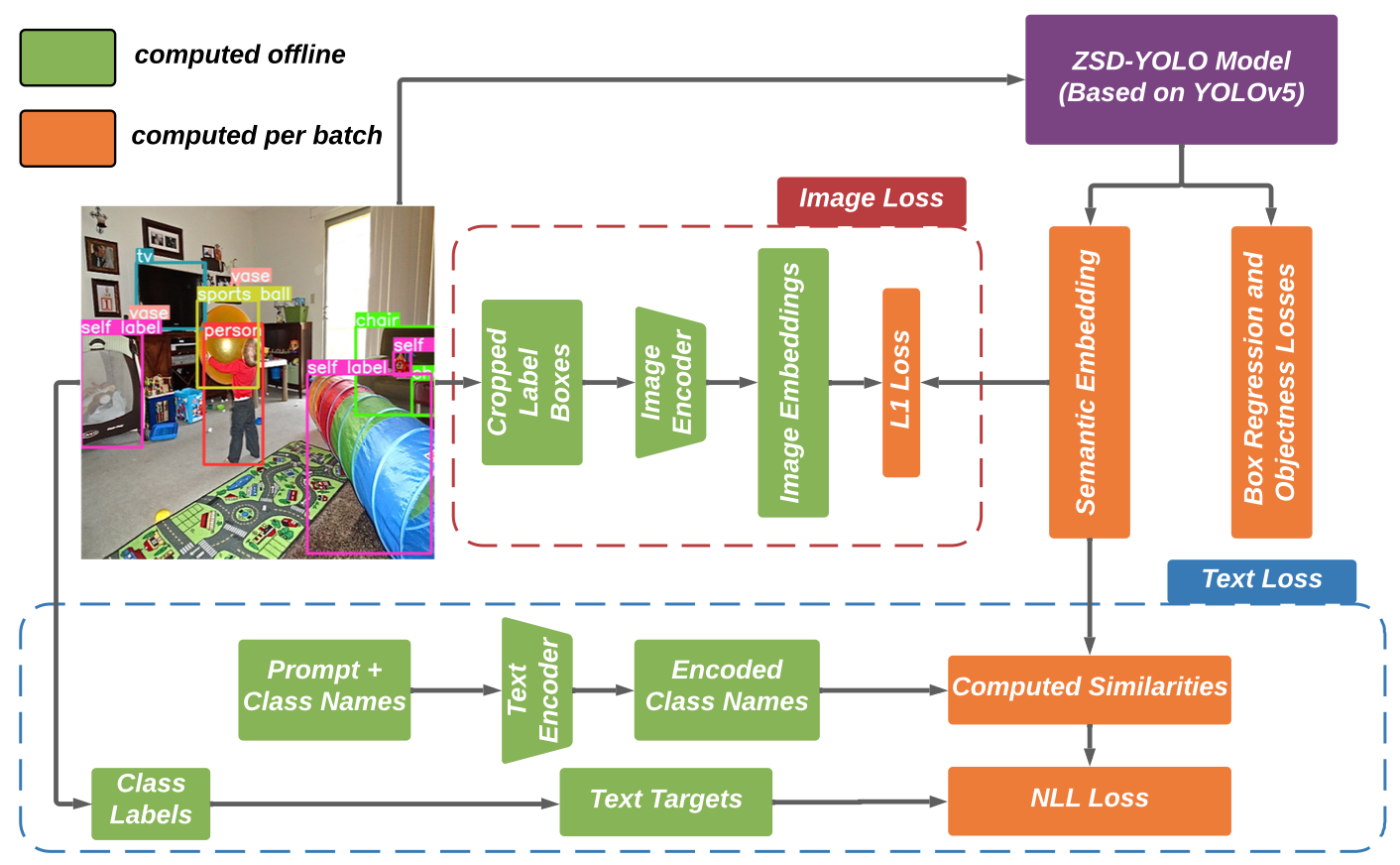}
    \caption{An overview of the proposed training method of ZSD-YOLO. The method aligns detector semantic outputs to vision and language embeddings from a pre-trained Vision-Language model such as CLIP. YOLOv5 has been modified to replace typical class outputs with a semantic output with a shape equal to the CLIP model embedding size and then align predicted semantic outputs of positively matched anchors with corresponding ground truth text embeddings with a modified cross-entropy loss. Image Embeddings of positively matched anchors are aligned using a modified L1 loss function. Best viewed in color\cite{xie2022zero}.}
    \label{figure1}
    
\end{figure}

In the domain of zero-shot object detection, research has pushed the boundaries of traditional object detectors by enabling them to recognize and locate objects even in entirely untrained categories. This novel paradigm leverages semantic alignment, aligning detector outputs with embeddings from pre-trained vision-language models, and has resulted in groundbreaking achievements such as ZSD-YOLO. Furthermore, self-labeling data augmentation methods have been developed to enhance model performance without the need for extra images or labels, along with the flexibility to apply the ZSD-YOLO model with varying network sizes under different computational constraints \cite{xie2022zero}, \cite{wu2023zero}. See figure \ref{figure1} for more information regarding the neural network structure. 

Zero-shot object detection extends its potential to medical imaging, where the exploration of visual-language pre-trained models (VLPMS) unlocks new horizons in unsupervised paradigms. This approach capitalizes on the rich semantic information learned from text-image pairs collected from the internet, enabling direct prediction of nuclei detection in histopathological images, a task previously underexplored. By combining the power of VLPMS with innovative design principles, a zero-shot label-free nuclei detection framework is established, outperforming existing unsupervised methods while ensuring compatibility with both seen and unseen objects \cite{wu2023zero}.  

Within the scope of zero-shot instance segmentation, where challenges lie in segmenting object instances of unseen classes and distinguishing between foreground and background, new methods emerge to address these issues. A novel approach tackles the problem in a background-aware detection-segmentation manner, introducing the Zero-shot Detector, Semantic Mask Network, Background Aware RPN, and Synchronized Background Strategy. Furthermore, a new experimental approach is introduced to evaluate the model's performance, accompanied by extensive experiments, highlighting the effectiveness of the proposed methods in surpassing the state-of-the-art in zero-shot detection and achieving promising outcomes in zero-shot instance segmentation \cite{zheng2021zero}, \cite{an2023zbs}, \cite{zheng2020background}.  

Apart from the previous way of thinking, a unique perspective comes from the research on predictions’ self-consistency over different text prompts for zero-shot classifiers, where an algorithm is designed to select the optimal prompts, maximizing model performance in a zero-shot fashion without relying on labeled validation data. Another proposal used Canonical Correlation Analysis (CCA) to improve the detection and localization of text in addition to a different model that enhances the phrases detection and reconstruction with the ability to deal with supervised and unsupervised data cases \cite{wu2023zero}, \cite{ge2023improving}, \cite{allingham2023simple}, \cite{rohrbach2016grounding}.  

Few-shot object detection enters the stage as a valuable connection between fully supervised and zero-shot methods. It offers an efficient and effective solution, harnessing the power of deep learning and vision-language models while overcoming challenges like domain gaps and overfitting. The introduction of a few-shot object detector presents a two-stage approach, enhancing performance by segmenting objects before searching and ensuring precision in object recognition. 

As we explore deeper into few-shot classification, we explore the synergy between large vision-language models and few-shot learning. By finetuning pre-trained models with a minimal number of examples, we harness the full potential of these models, demonstrating robustness and generalization capabilities that outperform traditional fine-tuning techniques. It also leverages semantic category labels, often overlooked, to facilitate more efficient few-shot learning, thereby achieving superior performance with limited examples \cite{xiao2022exploiting}.

Transitioning into the field of regular object detection, we encounter the vital role of deep learning in medical applications. Deep neural networks, particularly those equipped with convolutional neural network (CNN) backbones, have proven instrumental in the precise identification of cancer lesions and metastatic lymph nodes. These AI algorithms offer accurate lesion recognition, marking a significant step toward improved diagnostics \cite{yang2021tracking}.  

In addition, the application of object detection to medical images presents unique challenges, characterized by small target sizes, low image clarity, and substantial noise. Traditional object detection algorithms face difficulties in this context, emphasizing the need for tailored solutions. Researchers have introduced innovative techniques, such as the mask mechanism, to enhance the noise immunity and segmentation accuracy of medical images. In image Reconstruction phase, the mean square error (MSE Loss) is used to measure the difference between each pixel in the original image and the reconstructed pixel. The MSE Loss function is as follows: 
$$\mathscr{L}_{\mathrm{MSE}}=\frac{1}{N} \sum_{i=1}^N\left(y_i-\widehat{y}_i\right)^2$$
where N represents the total number of pixels in each image, $\widehat{y}_i$ represents the predicted value of the ith pixel value in the image, and yi represents the true value of the ith pixel value in the image. For the bounding box predition, the intersection over union (IoU) is used as a loss function, which is the ratio of the area of the intersection region and the area of the union region between the real and predicted boxes. The formula is as follows:  
$$ IoULoss =-\ln \frac{\text { Intersection }\left(\text { box }_{\mathrm{gt}}, \text { box }_{\text {pre }}\right)}{\text { Union }\left(\text { box }_{\mathrm{gt}}, \text { box }_{\text {pre }}\right)}$$

where boxgt represents the real box, and boxpre represents the prediction box. In general, the smaller the IoULoss is, the closer the coordinates of the ground-truth box and the predicted box are.
\cite{shou2022object}.

Furthermore, the use of depth maps and deep learning techniques generated by transformer-based networks effectively addresses the scale drift problem in monocular systems or what is called scale recovery, providing competitive performance, and eliminating the need for additional sensors \cite{franccani2022dense}.

In the domain of medical image segmentation, the limitations of conventional U-Net models are addressed by incorporating various improvements. These innovations aim to enhance classification performance and detail preservation, ensuring that even small and challenging objects in medical images can be accurately identified \cite{chen2020triple}.

Finally, the intersection of deep learning and computer-aided diagnosis (CAD) emerges as a powerful tool in the prognosis of premalignant and malignant stages of diseases. In this context, convolutional neural networks (CNNs) play a pivotal role, providing a promising way for advancing early disease detection and prognosis in the medical field \cite{article}.

This survey collectively highlights the pursuit in object detection, reaching the diverse landscapes of few-shot, zero-shot, and regular object detection, with applications in fields as varied as computer vision, segmentation, medical imaging, natural images, and beyond. In this survey, we reviewed more than twenty-one papers that discussed the excellence of few-shot, zero-shot, and object detection with medical and natural images for segmenting and classifying different objects. We categorized the papers regarding the type of data used are they medical or natural, learning method, supervised learning, or unsupervised learning, and finally the model type as discriminative or generative. This classification is explained in tables \ref{table1}, \ref{table2}, and \ref{table3} for zero-shot, few-shot, and regular object detection techniques respectively.

\section{Object detection}
\subsection{Zero-shot learning (ZSL)}
In recent years, ZSL has been an active topic with several reviews and methods. ZSL focuses on learning to recognize the properties of objects to tackle the unseen data during the training phase which are unlabeled data. In addition to other pain points, problems are tackled via new methods based on FSL.

A method was proposed by Johnathan et al. \cite{xie2022zero} for the addition of a one-stage detector to implement ZSD by aligning detector semantic outputs to embeddings from a trained vision-language model. Not only is learning derived from text embedding alignment but image embedding alignment is also incorporated, distinguishing it from other methods. The inclusion of this image alignment in the proposed method's loss function leads to a significant improvement in model performance. Ground truth bounding boxes were cropped by them and passed through the CLIP image encoder. They were then aligned using an L1 loss function. Additionally, a new post-processing operation was created by the authors as part of the ZSD task for YOLO models, aiming to enhance the detection of unseen classes.

In the paper titled A zero-shot nuclei detection framework was established by Wu et al. \cite{wu2023zero} based on Visual-language pre-trained models VLPM. The aim was to examine the capability of VLPM to facilitate the direct prediction of nuclei detection through semantic-driven prompts, creating a concise, clear, and more efficient transferable unsupervised system for label-free nuclei detection. Initially, the VLPM network BLIP was employed by them to automatically generate attribute words describing the unseen nuclei objects. BLIP possesses the capability to generate automatic descriptions for images. Subsequently, these attribute words were integrated with medical nouns, forming detection prompts in the format "[shape][color][noun]." These prompts were then utilized as inputs to Grounded Language-Image Pre-training GLIP, a VLPM model, to achieve zero-shot detection of nuclei. Furthermore, to enhance the precision of preliminary boxes, a self-training framework was established, using the preliminary boxes generated by GLIP as pseudo-labels for the subsequent training of YOLOX. The iterative manner and self-training strategy employed in this framework resulted in a remarkable performance for label-free nuclei detection, surpassing other comparison methods.

A new problem setting, termed zero-shot instance segmentation ZSI, was introduced by Zheng et al. \cite{zheng2021zero}. This setting is designed not only to detect all unseen objects but also to precisely segment each unseen instance further. The ZSI task encompasses two primary challenges:
1.	The challenge of performing instance segmentation for unseen classes without the availability of data from seen classes. To address this, extra semantic knowledge contained in pre-trained word vectors was utilized to establish correlations between seen and unseen classes. The semantic word vector and image data of seen classes were employed to establish visual-semantic mapping relationships in a detection-segmentation manner, which were then transferred to unseen classes. To achieve this, the zero-shot detector and semantic Mask Head (SMH) were proposed to detect and segment each unseen instance.
2.	The persistent challenge of confusion between the background and unseen classes. As ZSI necessitates distinguishing between foreground and background, the unseen classes are classified as background. To address this issue, the authors introduced the Background Aware RPN (BA-RPN) and synchronized Background Strategy (Sync-bg). Both were devised to tackle the background problem, contributing to the establishment of a reasonable and dynamically adaptive word vector for the background class.

A new background subtraction method called Zero-shot Background Subtraction ZBS  was introduced by An et al. \cite{an2023zbs} The method aims to overcome issues faced by other methods, especially the challenge of accurately distinguishing the edges of foreground objects with pixel-level background models. It involves three key stages:
1.	All-Instance Detection: Any zero-shot detector can be used. Detic, a zero-shot object detection model, was employed to transform raw image pixels into structured instance representations, including categories, boxes, and masks.
2.	Background Modeling: An instance-level background model is built based on the motion information of instances. If an object is stationary, it's added to the background model.
3.	Foreground Instance Selection: The algorithm selects the output of the all-instance detector when a new frame is received.
Benefiting from a comprehensive use of instance information, the proposed method performs well in challenging scenarios like shadows, camera jitter, and night scenes. ZBS also minimizes the chance of mistakenly identifying noisy backgrounds as foreground objects. It can detect most real-world categories and even identify unseen foreground categories not predefined.

In the paper \cite{hayat2020synthesizing}, a novel feature synthesis module, guided by semantic space representations, constitutes the core of this approach. It can generate diverse and discriminative visual features for unseen classes. Exemplars in the feature space were generated, and they were employed to modify the projection vectors corresponding to unseen classes in the Faster-RCNN classification head.

In the pursuit of Improved Zero-Shot Detection (ZSD), a generative method is proposed by Samra et al. \cite{sarma2022resolving}   The objective is to resolve semantic confusion by employing a triplet loss during the training of the feature synthesizer. The approach involves utilizing Faster-RCNN as the backbone object detector, trained on images exclusively containing seen class objects. Fixed-size feature vectors for these objects are then employed to train a conditional Wasserstein GAN.
To address the challenge of mode collapse in conditional GANs and ensure diversity among synthesized features, a regularization term is incorporated. However, the conditional Wasserstein GAN only learns to synthesize image features conditioned upon class semantics and overlooks the dissimilarity between object classes during feature synthesis. Consequently, a triplet loss is introduced to facilitate the learning of discriminative features for semantically similar classes, resolving semantic confusion when utilizing these synthesized features in the detection pipeline. Additionally, the aim is to maintain consistency between the synthesized visual features and the semantics of the corresponding class. This is achieved by incorporating a cyclic consistency loss, enforcing synthesized visual features to reconstruct their semantics. The trained conditional Wasserstein GAN is then utilized to generate features for unseen classes, which in turn are used to update the classifier of the pre-trained Faster-RCNN. This empowers the detector to effectively detect unseen-class objects. The performance of this classifier is directly tied to the quality of synthesized features used for training, highlighting the impact of accounting for inter-class dissimilarity and visual-semantic consistency on detector performance.

In the paper, it is affirmed by Nie et al. \cite{nie2022node}  that contextual information among multi-objects plays a more significant role in zero-shot detection than in traditional object detection. The Graph has been demonstrated as a superior tool for modeling visual and semantic relevance in various tasks. To capitalize on such contextual information, a novel ZSD approach named Graph Aligning Network (GRAN) is proposed based on graph modeling and reasoning. Specifically, for graph modeling, a Visual-Semantic Relational Graph (VSRG) is devised to comprehensively utilize both visual and semantic relational information. This involves constructing a Visual Relational Graph (VRG) and a Semantic Relational Graph (SRG). The nodes in these graphs represent objects in the image and the semantic representations of classes, respectively, while the edges denote the relevance between nodes in each graph. For graph reasoning, modal translators are designed for these two graphs to transform the node states of different models into a common space for communication. To update the representation of individual nodes with information from other nodes, each node first determines which messages to send. Subsequently, it receives visual and semantic messages from other nodes on the VSRG  that are highly relevant.

\begin{figure*}
    \centering
    \includegraphics[width=1\linewidth]{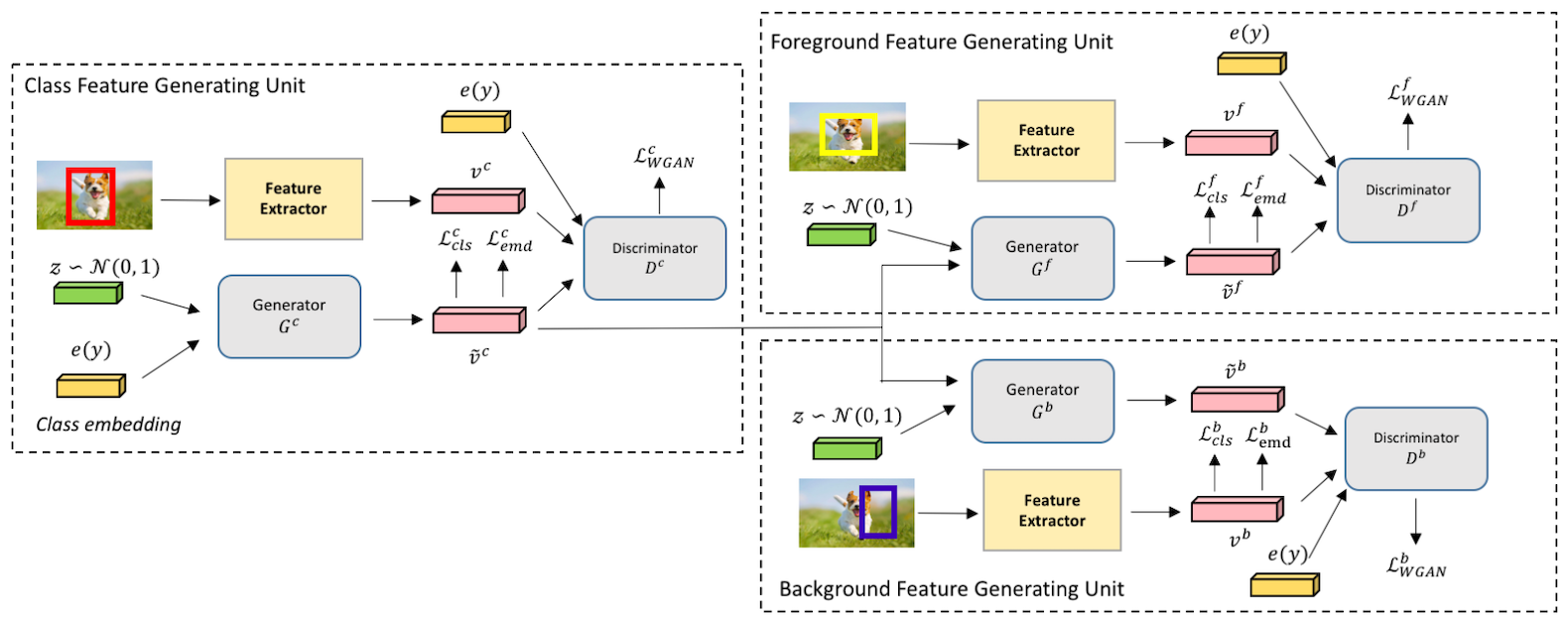}
    \caption{Illustration of the IoU-Aware Generative Adversarial Network (IoUGAN). The Class Feature Generating Unit (CFU) takes the class embeddings and the random noise vectors as input and outputs the features with the intra-class variance. Then the Foreground Feature Generating Unit (FFU) and the Background Feature Generating Unit (BFU) add the loU variance to the results of CFU and output the class-specific foreground and background features, respectively. \cite{zhao2020gtnet}}
    \label{figure2}
\end{figure*}

Another  text prompts based approaches have been explored as in ImageNet, many of the remaining failure cases were found by Ge et al. \cite{ge2023improving}  to be caused by noise and ambiguous text prompts related to the WordNet hierarchical structure of ImageNet. Some class names are sufficiently general, leading the model to struggle in correctly matching images from their specific subclasses. The analysis of failure modes suggests a high sensitivity of the text encoder to inputs, resulting in an overall lack of robustness in classification.
In addition to other observations, the proposal was made to first identify the subset of images for which the top-1 prediction is likely to be incorrect. Subsequently, an improvement in accuracy for those images was sought through a principled framework, augmenting their class labels by WordNet hierarchy. To estimate whether an image has an incorrect prediction, the consistency of predictions under different text prompt templates and image augmentations was used as a signal for prediction confidence estimation.
Commonly used prediction confidence scores, such as maximum SoftMax probability and maximum logit score, were found to be unreliable for the multi-modal CLIP and LiT models due to the poor calibration of the logit scores. Despite the unavailability of CLIP private training data, a hypothesis was formed suggesting that the common abbreviation "fig" for "figure" might be a contributing factor, occurring frequently in the training data and including many non-fruit illustrations.
In this work, a simple yet efficient zero-shot confidence estimation method was first proposed, better suited for CLIP. This method is based on the self-consistency of predictions over different text prompts and image perturbations. The idea, originally proposed by for improving the reasoning accuracy of large language models through self-consistency among multiple model outputs, was extended for confidence estimation in multi-modal models. The method proves effective at predicting mistakes.
The mapping of high-dimensional visual features to a low-dimensional semantic space often induces the hubness problem, attributed to the heterogeneity gap between these two spaces (Zhang and Peng 2018) \cite{zhang2018visual}. Addressing the hubness problem can be achieved by directly classifying an object in the visual feature space. Several zero-shot classification methods (Xian et al. 2018 \cite{xian2018feature}; Verma et al. 2018 \cite{verma2018generalized}; Li et al. 2019a \cite{li2019leveraging}; Huang et al. 2019 \cite{huang2019generative}) have demonstrated the effectiveness of this solution in the visual space. However, visual features encompass not only intra-class variance but also IoU variance, a crucial cue for object detection. To tackle these challenges, a Generative Transfer Network (GTNet) for ZSD is proposed by Zhao et al. \cite{zhao2020gtnet} Specifically, a generative model is introduced to synthesize visual features, addressing the hubness problem. Simultaneously considering the IoU variance, an IoU-Aware Generative Adversarial Network (IoUGAN) is designed to generate visual features incorporating both intra-class and IoU variances. The proposed GTNet comprises an Object Detection Module and a Knowledge Transfer Module. The Object Detection Module includes a feature extractor, a bounding box regressor, and a seen category classifier. The feature extractor is employed to extract features of the region of interest (RoI) from an image. The Knowledge Transfer Module consists of a feature synthesizer and an unseen category classifier. The feature synthesizer generates visual features for training the unseen category classifier. The trained unseen category classifier is then integrated with the feature extractor and bounding box regressor to achieve ZSD. IoUGAN consists of three unit models: Class Feature Generating Unit (CFU), Foreground Feature Generating Unit (FFU), and Background Feature Generating Unit (BFU). Each unit comprises a generator and a discriminator. CFU focuses on synthesizing features for each unseen class with intra-class variance conditioned on class semantic embeddings. FFU aims to add IoU variance to the CFU results, generating foreground features. Additionally, BFU synthesizes class-specific background features conditioned on the CFU results to reduce confusion between the background and unseen classes. Figure \ref{figure2} illustrates the IoU-Aware Generative Adversarial Network's three main units' structures.

Cao et al. \cite{cao2021gastric} Identifying gastric polyps accurately from the gastroscope poses a challenge. The gastric wall is covered with mucous membranes, leading to the formation of numerous folds. Some of these folds resemble gastric polyps, complicating the recognition of polyps. Additionally, certain gastric polyps have a small size, making them prone to being overlooked or misdiagnosed. The precise diagnosis of gastric polyps through gastroscopy remains challenging for doctors and even specialist examinations may result in a certain rate of missed detection. While computer-aided detection methods based on traditional approaches and deep learning methods have been employed, they prove insufficient in addressing the challenges and issues associated with this case. In response, a feature fusion and extraction module is proposed, integrated with the YOLOv3 network. In the traditional feature pyramid, new features can only fuse with corresponding level features on the backbone network and higher levels [39]. Each fusion operation dilutes feature information during transmission, especially from non-adjacent levels. In contrast to the traditional feature pyramid, their feature fusion and extraction module can simultaneously fuse different level features, generating a new feature pyramid that retains information from all levels. This module deepens the network, obtaining more semantic feature information crucial for distinguishing gastric polyps from gastric folds. The module effectively fuses low-level features with high-level features, enhancing the performance of gastric polyp detection. Built upon the YOLOv3 network, their network further improves object detection ability, especially for small objects.

Gouda et. al. \cite{gouda2023dounseen}. A must-have is an object detector capable of flexibly changing the subset of classes used. It would be more convenient and easier to possess a deep learning-based object detector without engaging in data collection or training. The distinction between deep learning-based template matching networks and zero-shot object detectors lies in the latter's ability to detect multiple object classes simultaneously. Unlike deep learning-based template matching, which aims to detect a single object class, zero-shot object detectors should have the capacity to detect numerous object classes concurrently. Detecting a single object implies linear time complexity for multiple objects, making deep learning-based template-matching networks somewhat less appealing due to their brute-force nature. Furthermore, zero-shot object detectors need to be cognizant of other objects in the environment that are not in the gallery set and cannot be classified. Recent research has predominantly focused on deep template matching. However, this work has been aimed to go beyond deep template matching by developing a zero-shot object detector for robotic grasping.

Zheng et. al\cite{zheng2020background},  Concerning the simultaneous localization and recognition of unseen objects, preliminary attempts exhibit limitations: (i) the inability to gradually optimize visual-semantic alignment for properly mapping visual features to semantic information; (ii) a lack of a convenient pipeline to learn a discriminative background class semantic embedding representation, crucial for reducing confusion between background and unseen classes; (iii) reliance on pre-trained weights learned from either seen or unseen datasets. As a solution, they proposed a novel framework named Background Learnable Cascade (BLC) for ZSD, comprising three components: Cascade Semantic R-CNN, semantic information flow, and BLRPN. BLC is inspired by cognitive science, specifically how humans reason about objects through semantic information.
Humans have established an abstract visual-semantic mapping relationship for seen objects and transferred it to recognize unseen objects. Inspired by this, BLC develops a visual-semantic alignment substructure named the semantic branch to learn the visual-semantic relationship between images of seen objects and word vectors. This alignment is then transferred from seen classes to unseen classes for detecting unseen objects. To progressively refine the visual-semantic alignment, BLC introduces Cascade Semantic R-CNN by integrating the semantic branch into a multi-stage architecture based on Cascade R-CNN. This combination leverages the cascade structure and multi-stage refinement policy. In Cascade Semantic R-CNN, semantic branches in later stages only benefit from better-localized bounding boxes without direct semantic information connections. To address this, BLC further designs the semantic information flow structure to enhance semantic information flow by directly connecting semantic branches in each cascade stage. The semantic feature in the current stage is modulated through fully connected layers and fed to the next stage. This design facilitates the circulation of semantic information between each stage, contributing to learning a proper visual-semantic relationship.
Due to the inadequacy of the coarse word vector for the background class used in the semantic branch to precisely represent the complex background, BLC introduces a novel framework called BLRPN to learn an appropriate word vector for the background class. The study demonstrates that replacing the coarse background word vector in the semantic branch with the new one learned from BLRPN effectively increases the recall rate for unseen classes.

Rethinking about prompts, Allingham et. al. \cite{allingham2023simple} highlighted the hand-crafted prompts' drawbacks. As regrettably, the necessity for a set of hand-crafted prompts to achieve satisfactory zero-shot performance significantly diminishes the promised general applicability of such zero-shot classifiers. Designing different sets of hand-crafted prompts can be labor-intensive, and the common prompt design processes demand access to a labeled validation dataset, which may not be available in practice. In this paper, the question was posed: "Can prompt engineering for zero-shot classifiers be automated?". Specifically, given a zero-shot model and a substantial pool of potential prompts, the goal is to select the optimal subset of prompts that maximize the model performance in a zero-shot fashion, i.e., without access to a labeled validation set.
The contributions are as follows:
1.	An algorithm is presented for automatically scoring the importance of prompts in a large pool concerning a specific downstream task when using text-image models for zero-shot classification. Subsequently, a weighted average prompt ensembling method is proposed, utilizing the scores as weights.
2.	Several pathologies are identified in a naive prompt scoring method, where the score can become overly confident due to biases in both pre-training and test data. These pathologies are addressed through bias correction in a zero-shot and optimization-free manner.
3.	The algorithm's performance is demonstrated to be superior to the existing approach of hand-crafted prompts, without the need for a labeled validation set and a labor-intensive manual tuning process. 

In the end Zero-shot learning (ZSL) enables models to recognize and classify unseen data during training, expanding the applicability of machine learning without the need for extensive retraining. On the other hand, ZSL faces challenges in model generalization and performance, especially with domain shifts and complex datasets, relying heavily on semantic embeddings and requiring careful parameter tuning.  The taxonomy of Zero-shot learning techniques can be visulalized in table \ref{table1}.

\begin{table*}[htbp]
    \centering
    \begin{tabular}{|p{1.5cm}| p{8cm} |p{1cm}| p{3.5cm} |p{2cm}|}\hline
    \textbf{Author's Name} & \textbf{Data used} & \textbf{Data type}  & \textbf{Probabilistic model type in ML} & \textbf{ML paradigms}\\\hline
        Johnathan et al. \cite{xie2022zero}  & COCO \cite{lin2014microsoft} dataset under the 65/15 and 48/17 ZSD class splits. For testing, they used the ZSD class split for ILSVRC \cite{russakovsky2015imagenet}and the visual Genome \cite{krishna2017visual} class split. & natural & discrmiminative & unsupervised \\ \hline
         
        Wu et al. \cite{wu2023zero}  & MoNuSeg dataset \cite{kumar2017dataset} with an average of 658 nuclei per image. The data split into training/testing of 16/14. & medical & discriminative & unsupervised \\  \hline
        
       Zheng et al.  \cite{zheng2021zero} & MS-COCO 2014 \cite{lin2014microsoft} 2014 as the basic dataset. For constructing the two division methods of seen and unseen data classes: 48/17 split and 65/15 split  & natural & discriminative & unsupervised \\   \hline
       
           An et. al. \cite{an2023zbs} & CDnet 2014 dataset \cite{wang2014cdnet}including 53 video sequences and 11 categories. & natural & discriminative & semi-supervised \\   \hline
           
           Hayat et. al. \cite{hayat2020synthesizing} & object detection datasets: MSCOCO 2014 \cite{lin2014microsoft}, ILSVRC Detection 2017 \cite{russakovsky2015imagenet},  and PASCAL VOC 2007/2012 \cite{everingham2010pascal}. For MSCOCO, with split 65/15 seen/unseen. & natural &  generative & unsupervised \\   \hline

           Sarma et. al. \cite{sarma2022resolving}   & two datasets in ZSD – MSCOCO \cite{lin2014microsoft} and PASCAL-VOC \cite{everingham2010pascal}. MSCOCO with seen/unseen split of 65/15. PASCAL-VOC 2007/2012 with 16/4 split. & natural & Generative  & supervised \\   \hline
           
           Nie et. al. \cite{nie2022node} & The validation on MSCOCO dataset \cite{lin2014microsoft}, which includes 82,783 training images and 40,504 validation images of 80 classes. Following the dataset splits of MSCOCO proposed, both splits of the dataset used: (1) 48/17 seen/unseen classes; (2) 65/15 seen/unseen classes.  & natural & discriminative & supervised \\   \hline
           
           Ge et. al. \cite{ge2023improving} & five different ImageNet- based datasets. & More likely to be natural & Discriminative based on confidence estimation and label augmentation & supervised \\  \hline
           
           Zhao et. al. \cite{zhao2020gtnet} & three datasets. ILSVRC-2017 \cite{russakovsky2015imagenet}, MSCOCO \cite{lin2014microsoft} and VisualGenome (VG) \cite{krishna2017visual}& Natural & Generative & supervised \\   \hline
           
           Cao et. al., \cite{cao2021gastric} & two datasets, which are the private gastric polyps dataset and the public colonic polyps dataset. The gastric polyps datasets A total of 2270 images. & medical & discriminative & supervised \\  \hline

           Gouda et. al. \cite{gouda2023dounseen} & two possibilities for datasets to train the zero-shot classifier. The first DoPose \cite{gouda2022category} and HOPE \cite{tyree20226} for testing and validation. The second possibility is the FewSol dataset \cite{jaykumar2022fewsol} for training. & NA &  discriminative & supervised \\  \hline
           
           Zheng et. al, \cite{zheng2020background} & MS-COCO dataset used. MS-COCO (2014) includes 82783 training images and 40504 validation images. MS-COCO with two different seen/unseen splits: (i) 48/17. (ii) 65/15.  & natural & discriminative &  supervised \\    \hline
           
           Allingham et. al., \cite{allingham2023simple} & Different sets of prompts were manually designed and tuned for different downstream tasks for CLIP. & natural & discriminative & supervised \\    \hline
    \end{tabular}
    \caption{datasets used for zero-shot and their classification.}
    \label{table1}
\end{table*}

\subsection{Few-shot learning (FSL)}
Different motivations led to more research in the field of FSL. For instance, the wide range of sizes of the Gastric polyps and the difficulty of the detection in gastroscopic images, many traditional methods were applied but still in some cases the detection is not accurate. On the other hand, DL methods gave much better results in detecting colonic polyps because of their capabilities. 
Chanting et al. \cite{cao2021gastric}. A method was developed to utilize a dataset of gastric polyps. The polyps were commonly located in the middle of the image. Detection methods employing Deep Learning (DL) can be categorized into two types – two-stage and single-stage detectors like YOLO. However, these methods exhibit a deficiency in detecting small objects due to the absence of detailed texture information. Consequently, these methods struggle to perform well on the gastric polyp dataset. To address this limitation, a feature extraction and fusion module were developed and integrated with the YOLOv3 network. The features were selectively combined from compatible levels on the backbone network and higher. Feature extraction facilitates the combination of diverse features, creating a new feature pyramid while retaining information from all levels. The feature fusion and extraction deepen the network, acquiring more information. Through the integration of low and high-level features, the performance of gastric polyp detection, especially for small objects, was enhanced.

Atsuyuki Miyai et al. \cite{miyai2023locoop}. For out-of-distribution (OOD), various studies have been conducted, including zero-shot methods and fully supervised methods. Zero-shot methods, not requiring training data, resulted in a domain gap that constrained their performance, while fully supervised methods necessitated some fine-tuning, impacting their effectiveness. Motivated by these limitations, the FS-OOD method that leverages the advantages of both zero-shot and fully supervised methods was developed. A CLIP was trained with only a few ID images using the prompt learning approach, wherein prompts were trained while keeping the pre trained context fixed. While this approach is typically achieved by CoOP, in the case of OOD, the results were unsatisfactory. Therefore, a new approach called Local Regularized Context Optimization (LoCoOp) was introduced. It was found that local features harbored ID-irrelevant nuisances, which LoCoOp learned and subsequently removed from the ID class to mitigate the production of high ID confidence scores, preventing the inclusion of uninteresting information. This process involved identifying irrelevant areas through classifications and then applying entropy maximization to the classification of these areas. This ensured that features in ID-irrelevant areas differed from any ID text embedding. The benefits of LoCoOp were twofold: first, the cost was low since detection trained the external OOD samples; second, LoCoOp utilized local features, aiding in accurately identifying whether an area is ID or OOD. The results indicated that LoCoOp exhibited significant improvements compared to existing zero-shot, few-shot, and fully fine-tuned OOD detection methods with only one label per image.

Taihong Xiao et al. \cite{xiao2022exploiting}, the motivation for this study was the issue in the new light of large vision-language models, where it was discovered that pre-trained models, such as ImageNet, can be used simply by fine-tuning on a different task. Similarly, the vision encoder from the pre-trained vision-language model can be fine-tuned and get into action. However, the vision-language models outperform ImageNet models because they are trained with significantly higher images and texts. It is noteworthy that larger vision-language models can lead to overfitting on limited data. Consequently, a new source of information was developed, derived from the category names in downstream image classification tasks. The aim was to demonstrate that this new approach would help vision-language models be transferred more effectively with few examples in downstream tasks. In this study, various scenarios for initializing a classification head were explored. The first scenario involves random initialization, where only the category ID is known without information about the category's meaning. The second scenario initializes the classification head with category names, extracting information from labels like "tench" and "goldfish." This process leverages a pre-trained language model to enhance model adaptation. The third scenario goes beyond English category names, considering class digits or non-English labels. Unlike the first scenario, which lacks text/language information, scenarios 2 and 3 use a pre-trained language model to parse text from categories. The baseline scenario (1) initializes the backbone network with pre-trained model weights and a randomly initialized classification head. For scenarios 2 and 3, the pre-trained language model processes category names and pairs them with prompts, extracting average text embeddings to initialize the classification head. Among these scenarios, category name initialization (CNI) in the second scenario demonstrates the best performance during fine-tuning with one-shot ImageNet data.

Few-shot learning (FSL) excels in enhancing object detection performance for specific challenges in such as gastric polyps in medical images, through the integration of feature extraction and fusion modules. But FSL methods may struggle with effectively handling out-of-distribution (OOD) data and achieving optimal performance and generalization, particularly in complex datasets and diverse class scenarios. Table \ref{table2} demonstrate the classification and dataset used in this section.

\begin{table*}[htbp]
    \centering
    \begin{tabular}{|p{1.5cm} |p{9.5cm} |p{1cm}| p{2cm} |p{2cm}|}\hline
    \textbf{Author's Name} & \textbf{Data used} & \textbf{Data type}  & \textbf{Probabilistic model type in ML} & \textbf{ML paradigms}\\\hline
        chanting et al., \cite{cao2021gastric} & gastric polyps’ dataset, which contained small and large polyps and the image had more than two gastric polyps. & medical &  discriminative & supervised \\  \hline
           
           Atsuyuki Miyai et al. \cite{miyai2023locoop} & ImageNet-1K dataset was used as the ID data. For OOD datasets, adoption of the same ones as in \cite{deng2009imagenet}, including subsets of iNaturalist \cite{van2018inaturalist}, SUN \cite{xiao2010sun}, Places \cite{zhou2017places}, and TEXTURE \cite{cimpoi2014describing}. & Medical & Discriminative & supervised \\    \hline
           
           Taihong Xiao et al. \cite{xiao2022exploiting} & Several datasets, such as ImageNet \cite{deng2009imagenet}, ImageNet-V2 \cite{recht2019imagenet}, ImageNet-R \cite{hendrycks2021many}, Oxford Flowers \cite{nilsback2008automated}, Stanford Cars \cite{krause20133d}, Country-211 \cite{radford2021learning},  EuroSAT \cite{helber2019eurosat}, and Oxford-IIIT Pets \cite{parkhi2012cats} & natural & discriminative & supervised \\  \hline

    \end{tabular}
    \caption{Few-shot learning datasets and their classification.} \label{table2}
\end{table*}

\subsection{Regular Object detection}
Bryan Plummer et al. \cite{plummer2020revisiting}. Introduced as a phrase detection benchmark without simplifications or restrictions, aiming to identify regions in images related to phrases within a test image database. Detection was measured by the percentage of accuracy of queries detected in images. The model provided a value representing the probability that a certain phrase was related to a specific region. However, a desirable localizer might be a fragile detector, as phrase localization often boiled down to discerning the fundamental object category of a phrase. Consequently, training models for localization led to overfitting, prompting the exploration of simpler methods like Canonical Correlation Analysis (CCA), which outperformed state-of-the-art approaches in phrase localization. They tested several methods, including CCA, the similarity network, and Query-Adaptive R-CNN. Significant differences were observed with CCA, which can be interpreted as whitening and aligning image regions and text features by considering the entirety of the dataset. Training neural networks with minibatches posed challenges in discriminating similar phrases as they only saw a fraction of the dataset in each minibatch. The conclusion drawn was that CCA could be applied to vision-language tasks by having it generate the layers of a neural network responsible for mapping visual and textual representations together, optimizing the CCA weights to enhance distinguishability. To enhance CCA performance in challenging image-language tasks, they considered approaches such as using WordNet for positive phrase augmentation (PPA). This method categorized a person in an area labeled as having a construction worker as positive, even if not explicitly marked as one, addressing sparse annotations related to grounding phrases. Overfitting was reduced by inverse frequency sampling (IFS), biasing minibatches to reduce the selection of common phrases during training. In summary, they argued that identifying phrases in an image without assuming their presence is a more challenging and meaningful task than localizing known phrases. Surprisingly, leading localization models excelled in localization but struggled in detection. In contrast, seemingly simpler CCA baselines performed better at distinguishing similar phrases and predicting phrase presence. Fine-tuning a CCA-initialized model yielded their best detection performance, suggesting that CCA serves as a fundamental data alignment or normalization process, enhancing cross-modal task performance.

On the same path of localization and detection of a phrase in given images, the latter scenario was considered by Anna Rohrbach et al. \cite{rohrbach2016grounding}.  The challenge of grounding arbitrary natural language phrases in images has been aimed to be addressed. For example, Localization annotations are generally lacking in the majority of parallel corpora containing sentence/visual data and are computationally expensive. Thus, their attended approach was localizing phrases based on images without bounding box annotations, also combining them with bounding box supervision when possible. A model was developed to reconstruct phrases based on the bounded box without the need of supervised data and image phrase localization. When this step is fulfilled, the model can forecast the right phrase. This can be done without supervision from additional bounding boxes; they named their method GroundeR as their model grounds and reconstructs. Additional supervision was added to the model by incorporating a loss function that penalizes incorrect attention before the reconstruction step. Finally, the GroundeR method was tested on the Flickr 30k Entities and ReferItGame datasets. Surprisingly, their unsupervised version outperformed previous methods, and the supervised approach beat the current best results on both datasets. What's intriguing is that the semi-supervised approach effectively used limited labeled data and outperformed the supervised version by leveraging multiple loss functions.

In monocular visual odometry, the dimensionality of the information poses a challenge and relies on a prior known absolute reference. Combining it with other sensors such as GPS, wheel odometry, or an inertial measurement unit (IMU) provides a reference scale.André O. Franc¸ani et al. \cite{franccani2022dense} stated that local optimization can be a conventional solution, like bundle adjustment (BA), loop closure (LC), and many other approaches. Meanwhile, significant differences were shown when applying DL to estimate dense depth maps from single images. This benefit was exploited by the researchers to address the scale recovery problem using a transformer-based network. The result showed that this approach demonstrated competitive performance compared to state-of-the-art solutions on the KITTI odometry benchmark.

Continuing in the exploring the medical challenges,another approach focused on gastric cancer for faster and earlier diagnosis, the possibility and efficiency of treatment can be increased. To achieve this, AI algorithms of object detection and semantic segmentation were employed by Ruixin Yang et al. \cite{yang2021tracking}. Constructing object detection and semantic segmentation algorithms with CNN backbones enabled the correct detection of lymph nodes. The results indicated that AI effectively detected cancer and predicted lesions in surrounding lymph nodes. It was claimed that this marked the first instance of AI being utilized for recognizing cancers in post-operative specimens.

Yuntao Shou et al. participated in tackling a medical challenge regarding the detection of lesions\cite{shou2022object}. Daily, physicians detect lesions, and with a large number of images, it becomes laborious and intensive. This can lead to visual fatigue and incorrect diagnoses. To solve this problem, DL was employed for automatic diagnosis. Nevertheless, this approach is still challenging because, in most cases, lesions are small, and the images still have noise. For instance, but not confined to, Girshick et al \cite{6909475}, used R-CNN to extract features and SVM for classification. Dosovitskiy et al. Furthermore in the same context of lesions' detection, the paper by Alexey Dosovitskiy et. al \cite{Dosovitskiy2020AnII}, developed the Vision Transformer (ViT) to divide the image into patches. The transformed features were input so the model could distinguish the image’s features using a self-attention mechanism. These methods and others can be used for general object detection, but when it comes to detecting small areas like lesions in medical imaging, which also have noises, these methods will not be efficient. One solution is performed by reducing the noise so that more rich features can be obtained for the detection, which can be achieved by the mask mechanism. In this paper, the transformer developed by Liu et al \cite{9878517} was optimized and combined with a hierarchical transformer introducing a self-attention mechanism called MS Transformer. However, consideration for images with low resolution was not taken into account by the model. The image was divided into patches (16 $\times$ 16) and then randomly sampled. The sampled patches were masked, refraining from participating in feature extraction during the encoding stage. Encoding and decoding were performed using Vision Transformer, with the unmasked patches serving as input for the encoder. The features were mapped into the latent feature space, and these features, along with unmasked patches, were used as input for the decoding. Finally, the model reconstructed and filtered the image by reducing the disparity between the input image and the decoded image.
The hierarchical transformer implemented a non-overlapping Windows self-attention mechanism to learn the crucial differences between features by utilizing the feature vector with rich semantic information as input. The window attention mechanism assigned a weighting for each window’s features, with smaller regions receiving higher weight, aiding the model in focusing on detecting small areas. Testing the model on benchmark datasets demonstrated out-performance compared to existing models.

Kumar C A et al. \cite{article} issued a method for the detection of Esophageal Cancer stages, indicating changes in the texture pattern of the esophagus lining, can be detected and diagnosed as non-dysplastic (noncancerous), High-grade dysplasia (HGD), and Low-grade dysplasia (LGD). The higher the survival rate, the earlier the detection, and this can be achieved through the application of AI, ML, and DL. Supervised DL architectures were utilized by Kumar A C et al. for the detection, segmentation, and classification of these stages. A tool named CAD, employing CNN for this approach, was developed using DL. The identification of white matter (WM), gray matter (GM), and cerebrospinal fluid (CSF) in infant brains holds paramount importance for monitoring brain health, especially in the context of morphological changes in neurodevelopmental disorders. The challenge arises from the similar intensity levels in both T1-weighted and T2-weighted images, leading to difficult segmentation.

\begin{figure*}[h]
    \centering
    \includegraphics[width=1\linewidth]{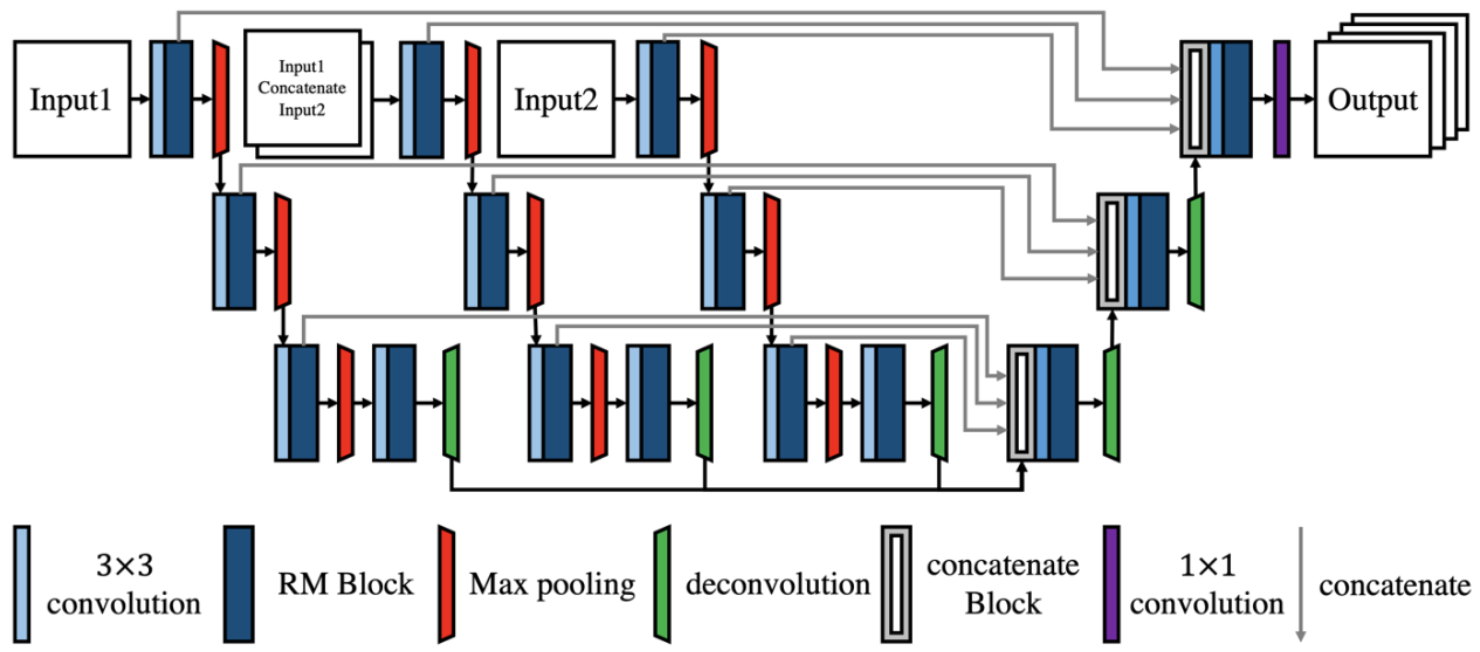}
    \caption{depicts the overall structure of the proposed model (TRMFCN) \cite{chen2020triple}, which is composed of encode and decode process. Encode process includes: traditional 2d convolution, maxpooling and residual multiscale (RM) block. Decode process includes: deconvolution, residuals multiscale (RM) block, concatenate block and traditional 2d convolution. The RM block is inspired by ResNet and Inception V1. The creation of this concatenate block is inherited from U-Net.}
    \label{figure3}
\end{figure*}

In the same context of segmentation, a model known as a Triple Residual Multiscale Fully Convolutional Network (TRMFCN) was developed by Yunjie Chen et al \cite{chen2020triple}. Using DL, the model is based on U-Net Ronneberger et al \cite{RFB15a}. The neural network in U-Net comprises two main components: a contracted path, primarily capturing image context through multiple 3×3 convolution layers and max-pooling, and an extended path, accurately locating the segmented image part through deconvolution and repeated 3×3 convolution layers. Skip connections enhance information utilization across scales. While extensive datasets are often required for deep learning networks, the U-Net effectively segments medical images with few samples, displaying good noise immunity. However, challenges such as weak edges in medical images and structural complexities, leading to increased parameters, are encountered. Enhancements have been proposed by numerous scholars, including DRINet with dense connections and block packaging, EDD+MUSCLE Net combining parallel architectures, ResUNet-a addressing gradient issues, MDU-Net with multi-scale dense structures, nested U-Net integrating feature information, SegNet optimizing with pooled index, PSPNet utilizing pyramid pooling, and DeepLab employing atrous convolution with a connected Condition Random Field (CRF). While these methods demonstrated success in various aspects, challenges like computational expense and resource requirements persist. To address the limitations observed in U-Net and its variant models, such as detail loss, prolonged training times, gradient disappearance, and slow convergence in image segmentation tasks, the TRMFCN model was introduced. This model incorporates three levels of input as shown in figure \ref{figure3}, effectively extracting information from multiple scales. Additionally, the Residual Multiscale (RM) block was introduced to facilitate easier convergence and employs the Concatenate Block to enhance information extraction. The incorporation of the residual multiscale block addresses the issue of gradient diffusion, enhancing the efficiency of network training. Furthermore, the concatenated block significantly improves feature reusability, enabling more comprehensive learning of global feature information. Notably, the model exhibits flexibility in handling NMR multi-modal image data, accommodating the seamless integration of new modal data by extending an additional branch, and the method is adaptable for single-mode MR image segmentation applications. 

Regular object detection techniques leverage advanced methods such as deep learning to accurately localize and identify objects within images. These techniques offer robust solutions applicable to various domains including medical imaging, visual odometry, and phrase detection. The reliance on large annotated datasets for training still can be time-consuming and expensive to acquire, especially for specialized domains. Additionally, these methods may struggle with detecting small or irregularly shaped objects, limiting their applicability in certain scenarios. Refer to table \ref{table3} for more detailed information regarding the data used in the papers.

\begin{table*}[htbp]
    \centering
    \begin{tabular}{|p{1.5cm} | p{7.5cm} |p{1cm}| p{2cm} |p{4cm}|}\hline
    \textbf{Author's Name} & \textbf{Data used} & \textbf{Data type}  & \textbf{Probabilistic model type in ML} & \textbf{ML paradigms}\\\hline
        Bryan Plummer et al. \cite{plummer2020revisiting}  & phrase localization methods were evaluated across three popular datasets: Flickr30K Entities consists of 276K bounding boxes in 32K images. ReferIt consists of 20K images. from the IAPR TC-12 dataset \cite{Grubinger2006TheIT} and Visual Genome \cite{Krishna2016VisualGC} consists of 108,077 images, which were divided into 77K/5K/5K images for training/testing/validation. & natural & discriminative & supervised \\    \hline

           Anna Rohrbach et al. \cite{rohrbach2016grounding} & Flickr 30k Entities \cite{7410660} and ReferItGame \cite{kazemzadeh-etal-2014-referitgame} contains over 99K regions from 20K images. & natural & Discriminative & Supervised, semi- and un- supervised \\ \hline
           
           Andr´e O. Franc¸ani et al. \cite{franccani2022dense}. & KITTI odometry dataset \cite{6248074}, consists of 22 sequences. & natural & discriminative & Supervised for DET depth maps and unsupervised for MVO overall monocular visual odometry \\   \hline
           
           Ruixin Yang et al. \cite{yang2021tracking} & 509 macroscopic images from 381 patients, and 57 macroscopic images from 48 patients were gathered for prospective verification. & medical & discriminative & supervised \\ \hline
           
            Yuntao Shou et al \cite{shou2022object} & DeepLesion \cite{article2}, the dataset contains 10,594 CT studies, and it contains 32,735 lesion annotations. Have been divided into training/validation/test sets with ratios of  70 percent, 15 percent, and 15 percent, and set the batch size to 32. & medical & discriminative & Unsupervised (selfsupervsied) \\   \hline
            
            Kumar C A  et al \cite{article} & Different methods with different datasets used & medical & discriminative & supervised \\  \hline
            
          Yunjie Chen et al \cite{chen2020triple} & 1-iSeg-2017 challenge (http://iseg2017.web.unc.edu), the average age of these babies collected was 6 months. 2-adult brain MR images from the Internet Brain Segmentation Repository, there are 2304 adult brain MR images. & medical & discriminative & supervised \\ \hline

    \end{tabular}
    \caption{Regular object detection datasets and their classification.}
    \label{table3}
\end{table*}

\section{Discussion and conclusion}
This scientific exploration provides a discussion of recent advancements in computer vision, specifically focusing on zero-shot learning, few-shot learning, and regular object detection with the common methods and challenges used in different papers such as addressing the misclassification of background as unseen class, small objects size and the use of text prompts in tackling different problems. The categorization of papers into these three domains reveals a common observation of shared challenges and varied solutions, motivating a comprehensive survey. Within the zero-shot learning domain, over thirteen papers were reviewed and summarized, highlighting key challenges related to bad classification and the complexity of predicting unseen classes. Innovative solutions, such as text and image embedding alignment, visual-language models, and the introduction of the "zero-shot instance segmentation" setting, were examined. Methods like Background-aware RPN, pixel-level background incorporation, and motion information utilization showcased effectiveness in addressing background-related issues and this is because of the good results observed.

Moreover, the Background Learnable Cascade for Zero-shot object detection introduced three crucial components—Cascade semantic R-CNN, semantic information flow, and BLRPN—to enhance model performance. Conditional GANs, Graph Aligning Networks (GRAN), and a self-consistency-based method were discussed as strategies to mitigate noise-related issues and improve reasoning accuracy. The mapping of high-dimensional visual features to a low-dimensional semantic space demonstrated significant impact, addressing the hubness problem and overcoming previous limitations. This discussion also covers challenges like ambiguous text prompts, presenting a zero-shot prompt weighting technique for improved prompt ensembling in text-image models that showed great potential.

In the field of few-shot learning, the challenge of detecting small objects lacking detailed texture information prompted the development of innovative methods.  The LoCoOp approach for few-shot out-of-distribution detection demonstrated accuracy improvements compared to existing methods. Motivated by the success of vision-language models over ImageNet models, a novel approach investigated scenarios for fine-tuning with a few examples, with the second scenario, initializing with category names, showing superior performance.

Moving into object detection, the method of Revisiting Image-Language Networks addressed overfitting during model training and the challenge of minibatches in discriminating similar phrases and this was a new good solution in our opinion to address such a problem. Canonical Correlation Analysis (CCA) was applied, and its performance was enhanced using WordNet for positive phrase augmentation (PPA). In medical applications, artificial intelligence plays a pivotal role in object detection and semantic segmentation for gastric cancer diagnosis, lymph node detection, and lesion diagnosis.

this scientific discussion highlights the ongoing development of computer vision, with advancements in zero-shot learning, few-shot learning, and object detection showcasing promising solutions to diverse challenges. These approaches hold great potential for real-world applications, particularly in medical imaging and generally in normal life tasks.

In summary, the various contributions highlighted in these papers drive the field of object detection into new domains, especially in the areas of few-shot, zero-shot, and regular object detection. The creation of innovative models like ZSD-YOLO, GTNet, and Cascade Semantic R-CNN, each designed to tackle specific challenges such as background misclassification, demonstrates a collective commitment to advancing the field. The introduction of novel techniques, such as background learnable cascade and zero-shot confidence scoring, indicates a devoted effort to enhance and optimize model performance.

The significance of these contributions is clear in metrics like mean Average Precision (mAP) and the capacity to address practical challenges in medical imaging, robotic grasping, and background subtraction. These proposed techniques not only boost the precision and adaptability of object detection models but also introduce advanced frameworks that induce a progressive shift in the field. However, it's crucial to recognize the limited discussion on development challenges, emphasizing the necessity for future research to explore the complexity of these state-of-the-art approaches.

Looking forward, the observed patterns point to a collective attempt to leverage vision-language models for diverse applications, prompting researchers to explore domain-specific adaptations and thoroughly explore limitations. Ultimately, these contributions help for a more robust, adaptable, and transformative aspect in object detection, fostering continued research and innovation at the intersection of computer vision and machine learning.

\bibliographystyle{ieeetr}
\bibliography{sources}

\section{Appendix}
\subsection{abbreviations}
\begin{description}
    \setlength{\itemsep}{0pt}
    \setlength{\labelsep}{0.1pt}
    \small
    \item[AI] Artificial Intelligence
    \item[BA] Bundle Adjustment
    \item[BA-RPN] Background Aware RPN
    \item[BFU] Background Feature Generating Unit
    \item[BLC] Background Learnable Cascade
    \item[BLRPN] Background Learnable Region Proposal Network
    \item[CCA] Canonical Correlation Analysis
    \item[CFU] Class Feature Generating Unit
    \item[CLIP] Contrastive Language-Image Pre-Training
    \item[DL] Deep Learning
    \item[FFU] Foreground Feature Generating Unit
    \item[GLIP] Grounded Language-Image Pre-training
    \item[GRAN] GRaph Aligning Network
    \item[GTNet] Generative Transfer Network
    \item[IFS] Inverse Frequency Sampling
    \item[IMU] Inertial Measurement Unit
    \item[IoUGAN] IoU-Aware Generative Adversarial Network
    \item[LC] Loop Closure
    \item[LoCoOp] Local Regularized Context Optimization
    \item[OOD] Out-of-Distribution
    \item[PPA] Positive Phrase Augmentation
    \item[RoI] Region-of-Interest
    \item[SMH] Semantic Mask Head
    \item[SRG] Semantic Relational Graph
    \item[VRG] Visual Relational Graph
    \item[VSRG] Visual-Semantic Relational Graph
    \item[YOLO] You-Only-Look-Once
    \item[ZBS] Zero-Shot Object Detection
    \item[ZSD] Zero-Shot Detection
    \item[ZSI] Zero-Shot Instance Segmentation
    \item[ZSL] Zero-Shot Learning
\end{description}

 \end{document}